\pdfoutput=1
\documentclass{article}
\usepackage{iclr2027_conference,times}

\usepackage{amsmath,amsfonts,bm}

\def\eqref#1{equation~\ref{#1}}

\def\1{\bm{1}}

\DeclareMathAlphabet{\mathsfit}{\encodingdefault}{\sfdefault}{m}{sl}
\SetMathAlphabet{\mathsfit}{bold}{\encodingdefault}{\sfdefault}{bx}{n}

\usepackage{hyperref}
\usepackage{url}
\usepackage{booktabs}
\usepackage{tikz}
\usetikzlibrary{arrows.meta,decorations.pathreplacing}
\usepackage{xcolor}
\usepackage{enumitem}
\usepackage{graphicx}
\usepackage{multirow}
\usepackage{colortbl}
\providecolor{mycolor}{RGB}{214,228,240}
\usepackage{wrapfig}
\usepackage{xspace}

\newcommand{\ours}{Prefilling-dLLM~v2\xspace}

\title{Prefilling-dLLM v2: Learned Chunk Routing for Long-Context Diffusion Language Models}

\author{Qi Han$^{1}$, Jing Xiong$^{1,\dagger}$, Ziyuan Zhang$^{2}$, Ngai Wong$^{1}$ \\
$^{1}$University of Hong Kong \quad $^{2}$Zhejiang University \\
$^{\dagger}$Corresponding author}

\iclrfinalcopy
\begin{document}

\maketitle
\lhead{Preprint.}

\begin{abstract}
Diffusion language models apply bidirectional attention over the whole context at every denoising step, so a longer context makes each step substantially more expensive.
Sparsifying only at inference leaves the pretrained weights unchanged, so the model is never trained under the sparse pattern it runs with.
We propose \ours, which continues training a diffusion language model so that it routes with its native mask token.
One mask is appended at the end of each chunk, and a learned role offset marks it as a routing slot.
On every fourth layer, a query keeps a bidirectional local window and the top-$K$ chunks scored from that mask; the other layers use a bidirectional sliding window.
Starting from Dream-v0-Base-7B extended to 16K with YaRN, we compare \ours with bidirectional NSA, DSA, and SWA under the same layout, key budget, and dense initialization, using the same Fast-dLLM decoder.
With the YaRN factor scaled to the evaluation length, \ours leads the other sparse methods on all three RULER tasks from 16K to 128K.
Its mean accuracy at 128K is 7.4 points above NSA, the next sparse model, while DSA and SWA remain near zero.
At the same length, decoding is $172\times$ faster than with full dense attention.
\end{abstract}

\section{Introduction}

\begin{wrapfigure}[16]{r}{0.36\textwidth}
\vspace{-1.4em}
\centering
\includegraphics[width=\linewidth]{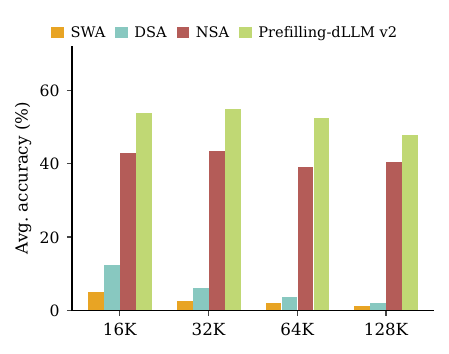}
\caption{Mean accuracy of the four sparse methods on the three RULER tasks.\ The YaRN factor scales with the evaluation length.}
\label{fig:ruler-sparse}
\end{wrapfigure}

Diffusion language models (dLLMs) generate text by iteratively unmasking
tokens under bidirectional attention, rather than by left-to-right next-token
prediction~\citep{austin2021d3pm,sahoo2024mdlm,nie2025llada,ye2025dream}.
At every denoising step, each query can attend to the whole context, so a
longer context makes that step substantially more expensive.

Sparse attention in autoregressive models allocates a limited budget with a local window, a learned token-level route, or a mixture of compressed and selected blocks~\citep{jiang2023mistral,yuan2025nsa,deepseek2025v32}.
These designs focus on causal attention, whereas efficient dLLM inference is studied mainly in a \emph{training-free regime}.
Fast-dLLM reuses keys and values across denoising steps and decodes blocks in
parallel; it accelerates serving but does not sparsify
attention~\citep{wu2025fastdllm}.
Sparse-dLLM evicts cache entries~\citep{song2025sparsedllm}, SparseD builds a
head-specific block-sparse mask at inference time~\citep{wang2025sparsed}, and
Prefilling-dLLM scores prefix chunks and reuses a static top-$K$
cache~\citep{xiong2026prefilling}.
Sparsifying only at inference restricts which tokens are visible and leaves
the pretrained dense attention weights unchanged, so the model runs under a
sparse pattern it was never trained with.
Training a dLLM under a sparse attention budget has received comparatively
less study.

\begin{figure*}[t]
\centering
\begin{tikzpicture}[x=1cm, y=1cm, font=\sffamily]
  \definecolor{Idle}{HTML}{F3F0EA}
  \definecolor{Win}{HTML}{88C8C0}
  \definecolor{Sel}{HTML}{E8A424}
  \definecolor{Mask}{HTML}{B45C58}
  \definecolor{Sum}{HTML}{8A8478}
  \definecolor{Ink}{HTML}{2C2C2C}
  \fill[Sum, rounded corners=0.8pt] (3.15,1.14) rectangle ++(0.62,0.10);
  \node[anchor=west, font=\scriptsize, text=Ink] at (3.88,1.18) {summary};
  \fill[Win, rounded corners=1.2pt] (6.15,1.02) rectangle ++(0.46,0.32);
  \node[anchor=west, font=\scriptsize, text=Ink] at (6.72,1.18) {local window};
  \fill[Sel, rounded corners=1.2pt] (9.05,1.02) rectangle ++(0.46,0.32);
  \node[anchor=west, font=\scriptsize, text=Ink] at (9.62,1.18) {selected};
  \foreach \side/\xzero in {0/0, 1/8.05} {
    \foreach \k in {0,1} {
      \foreach \j in {0,1,2} {
        \pgfmathsetmacro{\xx}{\xzero+\k*2.20+\j*0.70}
        \fill[Idle, rounded corners=1.5pt] (\xx,0) rectangle ++(0.62,0.56);
      }
    }
  }
  \foreach \j in {0,1,2} {
    \pgfmathsetmacro{\xx}{4.85+\j*0.70}
    \fill[Win, rounded corners=1.5pt] (\xx,0) rectangle ++(0.62,0.56);
  }
  \node[font=\scriptsize\bfseries, text=Ink] at (5.86,0.28) {$q$};
  \node[anchor=east, font=\small\bfseries, text=Ink] at (-0.22,0.28) {SWA};
  \foreach \side/\xzero in {0/0, 1/8.05} {
    \foreach \k in {0,1} {
      \pgfmathsetmacro{\bx}{\xzero+\k*2.20}
      \fill[Sum, rounded corners=0.8pt] (\bx,-0.50) rectangle ++(2.02,0.09);
      \foreach \j in {0,1,2} {
        \pgfmathsetmacro{\xx}{\bx+\j*0.70}
        \fill[Idle, rounded corners=1.5pt] (\xx,-1.22) rectangle ++(0.62,0.56);
      }
    }
  }
  \foreach \xzero in {0, 10.25} {
    \foreach \j in {0,1,2} {
      \pgfmathsetmacro{\xx}{\xzero+\j*0.70}
      \fill[Sel, rounded corners=1.5pt] (\xx,-1.22) rectangle ++(0.62,0.56);
    }
  }
  \foreach \j in {0,1,2} {
    \pgfmathsetmacro{\xx}{4.85+\j*0.70}
    \fill[Win, rounded corners=1.5pt] (\xx,-1.22) rectangle ++(0.62,0.56);
  }
  \node[font=\scriptsize\bfseries, text=Ink] at (5.86,-0.94) {$q$};
  \node[anchor=east, font=\small\bfseries, text=Ink] at (-0.22,-0.94) {NSA};
  \foreach \side/\xzero in {0/0, 1/8.05} {
    \foreach \k in {0,1} {
      \foreach \j in {0,1,2} {
        \pgfmathsetmacro{\xx}{\xzero+\k*2.20+\j*0.70}
        \fill[Idle, rounded corners=1.5pt] (\xx,-2.44) rectangle ++(0.62,0.56);
      }
    }
  }
  \foreach \xz/\j in {0/0, 2.20/2, 8.05/1, 10.25/0} {
    \pgfmathsetmacro{\xx}{\xz+\j*0.70}
    \fill[Sel, rounded corners=1.5pt] (\xx,-2.44) rectangle ++(0.62,0.56);
  }
  \foreach \j in {0,1,2} {
    \pgfmathsetmacro{\xx}{4.85+\j*0.70}
    \fill[Win, rounded corners=1.5pt] (\xx,-2.44) rectangle ++(0.62,0.56);
  }
  \node[font=\scriptsize\bfseries, text=Ink] at (5.86,-2.16) {$q$};
  \node[anchor=east, font=\small\bfseries, text=Ink] at (-0.22,-2.16) {DSA};
  \foreach \side/\xzero in {0/0, 1/8.05} {
    \foreach \k in {0,1} {
      \pgfmathsetmacro{\bx}{\xzero+\k*2.20}
      \foreach \j in {0,1} {
        \pgfmathsetmacro{\xx}{\bx+\j*0.70}
        \fill[Idle, rounded corners=1.5pt] (\xx,-3.66) rectangle ++(0.62,0.56);
      }
      \pgfmathsetmacro{\mx}{\bx+1.40}
      \fill[Mask, rounded corners=1.5pt] (\mx,-3.66) rectangle ++(0.62,0.56);
      \node[font=\tiny\bfseries, text=white] at (\mx+0.31,-3.38) {mask};
    }
  }
  \foreach \xzero in {0, 10.25} {
    \foreach \j in {0,1} {
      \pgfmathsetmacro{\xx}{\xzero+\j*0.70}
      \fill[Sel, rounded corners=1.5pt] (\xx,-3.66) rectangle ++(0.62,0.56);
    }
  }
  \foreach \j in {0,1,2} {
    \pgfmathsetmacro{\xx}{4.85+\j*0.70}
    \fill[Win, rounded corners=1.5pt] (\xx,-3.66) rectangle ++(0.62,0.56);
  }
  \node[font=\scriptsize\bfseries, text=Ink] at (5.86,-3.38) {$q$};
  \node[anchor=east, align=right, font=\scriptsize\bfseries, text=Ink] at (-0.22,-3.38) {Prefilling-\\dLLM v2};
  \foreach \yy in {0.28,-0.94,-2.16,-3.38} {
    \node[font=\scriptsize, text=Ink] at (4.54,\yy) {$\cdots$};
    \node[font=\scriptsize, text=Ink] at (7.46,\yy) {$\cdots$};
  }
\end{tikzpicture}
\caption{What query $q$ can see.
Each chunk is three cells, and $q$ sits in the middle so the local window extends both ways.
Teal cells are the local window and orange cells are selected.
The bar above each NSA chunk is its summary.
DSA reads scattered tokens, and \ours marks the last cell of every chunk as a mask.}
\label{fig:sparse-contracts}
\end{figure*}
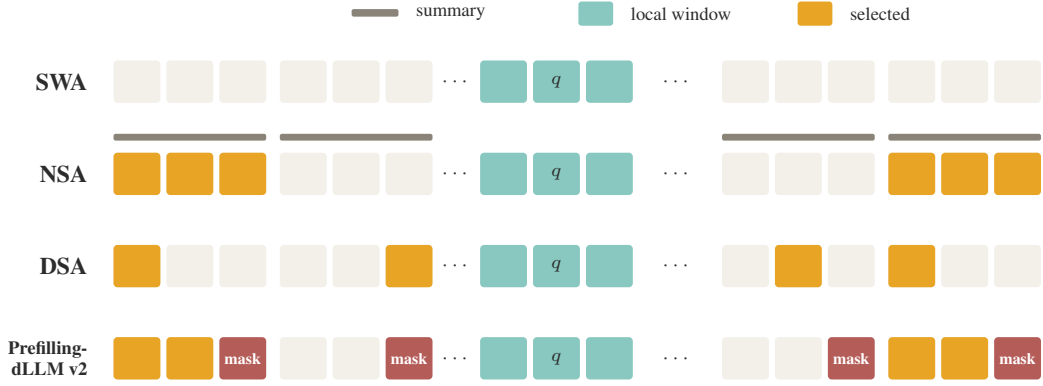

Our key insight is that a mask token is trained to absorb the context needed
to predict an unknown token.
Fixed at the end of a chunk, it can summarize that chunk and indicate which
remote chunks a query should read.
We therefore propose \textbf{\ours}, a sparse attention method learned by
\emph{continued training}, which selects remote chunks with these summaries.
Figure~\ref{fig:sparse-contracts} contrasts the resulting visibility with
bidirectional adaptations of Native Sparse Attention (NSA), DeepSeek Sparse
Attention (DSA), and sliding-window attention (SWA).

Specifically, \ours appends the model's native mask token at the end of each
chunk.
A binary indicator marks that position as a routing position rather than one
to be generated, and a learned role offset is added only there.
Through the shared query and key projections, the offset changes how the
routing mask attends, while a mask marked for generation remains a token to be
predicted.
The routing mask's query attends over the tokens in its chunk and pools their
keys into a summary.
The pooling weights are a softmax over the content keys, and a chunk score
adds the entropy of those weights to the similarity between the query and the
summary.
Each query keeps the highest-scoring chunks.
The same scores are then normalized together with the local branch, and the
resulting weights assign attention mass to each selected chunk and to the
bidirectional local window.
Each of the other layers applies a bidirectional sliding window around the
query.
Routing positions are omitted from the language-model targets and from the
value stream, so the summary is used to choose tokens rather than being
predicted or copied into the answer.

Starting from a Dream-v0-Base-7B checkpoint extended to 16K with
YaRN~\citep{peng2024yarn}, we compare \ours with NSA, DSA, and SWA under the same layer layout, key budget,
and dense initialization, using the same Fast-dLLM decoder at
evaluation~\citep{wu2025fastdllm}.
The benchmarks are LongBench-v1~\citep{bai2024longbench} and three RULER
tasks: single-needle retrieval, multi-key multi-query, and variable
tracking~\citep{hsieh2024ruler}.
Figure~\ref{fig:ruler-sparse} plots the mean accuracy on the three tasks
with the YaRN factor scaled to the evaluation length: \ours reaches 53.8 at
16K and 47.7 at 128K.
NSA, the next sparse model, scores 42.7 and 40.3, so the margins are 11.1 and
7.4 points.
DSA and SWA stay below 13 points at both lengths.
Single-needle accuracy is 100.0 at 16K and 95.0 at 128K, against 81.0 and
83.0 for NSA.
With the YaRN factor fixed at its training value, single-needle accuracy is
100.0 at 16K and 64K, and 99.0 at 32K and 128K.
Component and routing-token ablations are reported in
Appendix~\ref{sec:ablation} and are not part of this comparison.
Our contributions are as follows:
\begin{itemize}[leftmargin=*,itemsep=2pt,topsep=3pt]
  \item Our core contribution is \textbf{\ours}, a sparse attention method for
  dLLMs, learned by \emph{continued training}, that summarizes each chunk with
  a native mask token and trains the model to select remote chunks from those
  summaries.
  \item We introduce \emph{mask-summary scoring}, which marks that mask as
  a routing position with a binary indicator and a learned role offset, pools the chunk keys into a summary, and normalizes the scores with
  the local branch to assign attention mass.
  \item On three RULER tasks, with the YaRN factor scaled to the evaluation
  length, \ours leads the other sparse methods on every task from 16K to 128K.
  The three-task mean is 11.1 points above NSA at 16K and 7.4 points above NSA
  at 128K.
\end{itemize}

\section{Related Work}

\paragraph{Diffusion language models.}
Diffusion was first applied to text by corrupting continuous word embeddings and learning to reverse that process~\citep{li2022diffusionlm,gong2023diffseq}.
A second line corrupts discrete tokens.
Absorbing-state and masked diffusion replace selected tokens with a mask and train a model to reconstruct them under bidirectional context~\citep{austin2021d3pm,he2023diffusionbert,sahoo2024mdlm}.
This masked formulation has been scaled to LLM size along two routes.
One converts a pretrained autoregressive model into a diffusion model by continued training~\citep{gong2025adapt,ye2025dream}.
The other trains from scratch: LLaDA is competitive with autoregressive models at the 8B scale~\citep{nie2025llada}.
Later work aligns these models with preferences~\citep{zhu2025llada15}, adds visual instruction tuning~\citep{you2025lladav}, and extends the context window~\citep{liu2025longllada,peng2024yarn}.
Across these extensions, every denoising step still uses dense bidirectional attention over the whole sequence.

\paragraph{Efficient diffusion language models.}
One line reduces attention cost by training under a restricted mask.
Block diffusion denoises one block at a time: tokens inside a block attend bidirectionally, while each block attends only to earlier blocks~\citep{arriola2025bd3lm}.
Fast-dLLM v2 continues training from a pretrained autoregressive model under this mask, and caches completed blocks at inference~\citep{wu2025fastdllmv2}.
A second line leaves the dense bidirectional backbone unchanged and accelerates serving.
Fast-dLLM reuses key and value states across denoising steps and decodes a block in parallel~\citep{wu2025fastdllm}.
SparseD builds a head-specific block-sparse mask at inference~\citep{wang2025sparsed}, Sparse-dLLM evicts cache entries~\citep{song2025sparsedllm}, and Prefilling-dLLM prefills prefix chunks once, retrieves a top-$K$ subset, and reuses that cache for the remaining steps~\citep{xiong2026prefilling}.
The trained masks above are fixed in advance, and the serving methods decide visibility only after dense training.
Training a bidirectional dLLM to select remote context under a limited attention budget has received less study.

\paragraph{Sparse attention in autoregressive LMs.}
Sparse attention for autoregressive models falls into three groups: fixed patterns, patterns learned in training, and patterns applied only at inference.
Fixed patterns keep each query inside a local window, a stride, or a few global tokens~\citep{child2019sparsetransformer,beltagy2020longformer,zaheer2020bigbird,jiang2023mistral}.
A distant token is read only when it falls inside that pattern.
Learned patterns train the model under a limited budget: NSA mixes compressed, selected, and sliding branches~\citep{yuan2025nsa}, DSA selects individual tokens with a lightweight indexer~\citep{deepseek2025v32}, and MoBA and HiLS route each query to a few chunks~\citep{lu2025moba,hu2026hils}.
Serving-time patterns leave the causal weights unchanged and drop tokens through streaming caches, eviction, or dynamic block masks~\citep{xiao2024streamingllm,zhang2023h2o,xiong2025uncomp,jiang2024minference}.
Eviction can also be applied to chunks that are encoded in parallel, which extends the context length without training~\citep{xiong2025parallelcomp}.
These designs assume causal attention and a cache that grows from left to right.
Whether the same designs remain useful once attention is bidirectional, with the budget imposed while training a dLLM, is still open.

\section{Method}
\label{sec:method}

\begin{figure*}[t]
\centering
\includegraphics[width=\textwidth]{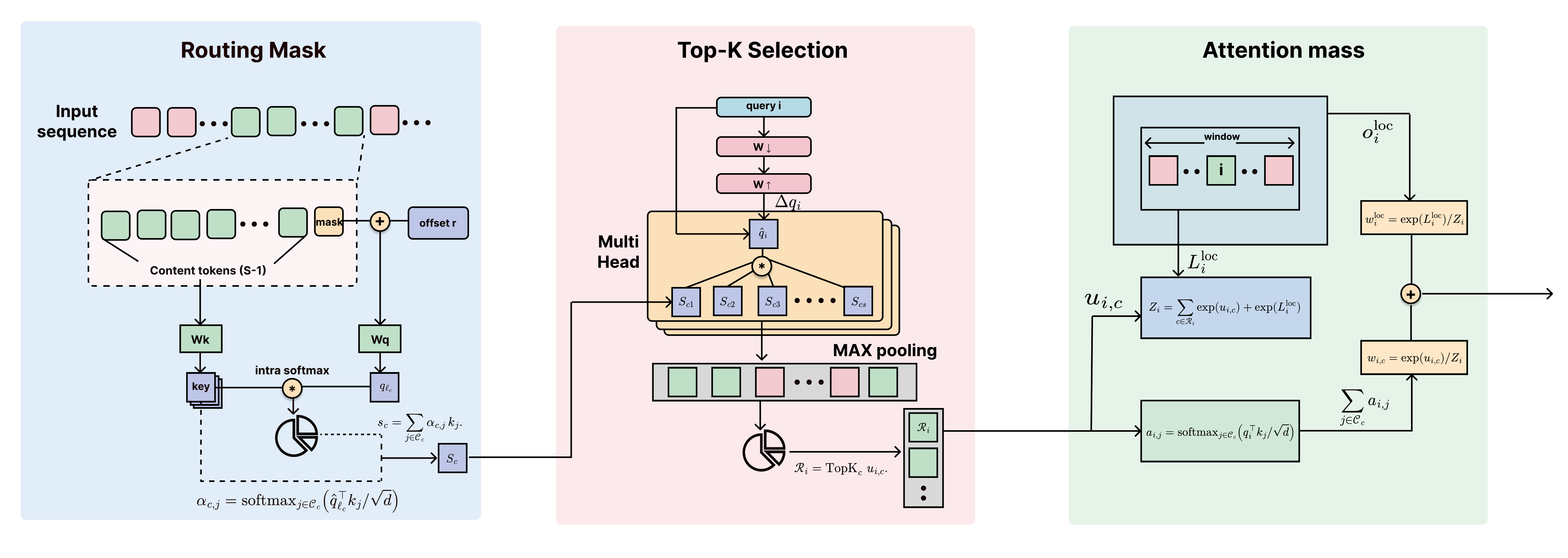}
\caption{Computation on a route layer.
A chunk is $S-1$ content tokens followed by one routing mask.
The role offset $r$ is added to that mask at the input.
Its query pools the content keys with an intra-chunk softmax, and the
weighted sum is the summary key $s_c$.
A content query adds a low-rank residual and scores these summaries.
Inside a GQA group~\citep{ainsworth2023gqa} the scores are reduced by a maximum, and the top $K$
chunks are shared by every head in the group.
Attention mass puts the selected chunk scores and the local-window
log-sum-exp in one partition, then a second softmax distributes each
chunk's mass over its content tokens.}
\label{fig:method}
\end{figure*}

\ours is a sparse attention method for a masked diffusion language model,
learned by continued training.
At one denoising step the model sees the input sequence $x_{0:L-1}$, which
includes the appended routing positions.
A query at position $i$ may attend to valid keys on either side of $i$.
Figure~\ref{fig:method} follows the computation on a route layer in three parts:
the routing mask that represents a chunk
(Section~\ref{sec:routing-mask}),
the top-$K$ selection of remote chunks
(Section~\ref{sec:topk-selection}),
and the attention mass those scores assign
(Section~\ref{sec:attn-mass}).
The other layers are non-route layers and use sliding-window attention.
A local window in this paper is centered on a query and extends to both
sides of it.
We give its size as a radius: $\rho$ on a non-route layer, and $w$ for the
local branch of a route layer.

The repeated $3\mathrm{SWA}+1\mathrm{route}$ layout follows from the
bidirectional information flow of a dLLM.
A query at an early position can attend to tokens that occur later in the
sequence.
If the layers between route layers were dense, that remote evidence could be
mixed into the hidden states before the next route decision, and the learned
route would no longer be the only long-range path.
Sliding-window layers keep bidirectional mixing inside the local radius and
block direct attention beyond that radius, so long-range evidence enters at
the route layers.
It can then be carried by the hidden states through the following local layers.
The window limits distance rather than imposing a causal direction.
The layout therefore serves two purposes: it bounds the cost, and it makes
the learned route the only long-range path.
The layer schedule, the radii, the chunk size, and the number of selected
chunks are fixed for the comparison and listed in
Appendix~\ref{sec:matched_protocol}.

\subsection{Routing mask}
\label{sec:routing-mask}

\paragraph{Chunk splitting.}
The sequence is partitioned into contiguous chunks of $S$ positions.
Chunk $c$ holds $S-1$ content tokens, followed by one routing position
$\ell_c$ at the end of the chunk.
The routing position is filled with the model's native mask id, so the vocabulary is unchanged
and the embedding $E$ of the mask is the one the model was already trained
with.
The same id is also written over content positions that the loss will predict.
A binary indicator $m_i\in\{0,1\}$ separates the two uses.
It is stored with the validity mask, and $m_i=1$ only at the appended routing
positions.
A mask that occupies a position to be denoised has $m_i=0$.

\paragraph{Role offset.}
The role offset $r$ is one learned vector of the model's hidden size,
shared by every chunk and every denoising step.
It is added once, before the first layer, and only on a routing position:
that position still stores the native mask id, and the input to the first
layer is the mask embedding plus $r$,
\begin{equation}
  e_i = E(x_i) + m_i r.
  \label{eq:role-offset}
\end{equation}
A content position has $m_i=0$ and keeps $E(x_i)$, and no later layer adds
$r$ again.
The first three layers are sliding-window layers.
The first route layer is the fourth layer, and the remaining route layers
follow every four layers.
A route layer does not read $e_i$ directly.
Let $h_i$ be the hidden state that reaches the layer, and let $R_i$ be the
rotary map at position $i$.
The backbone query is the hidden state under the shared projection,
\begin{equation}
  q_i = R_i\!\left(W_Q h_i\right).
  \label{eq:shared-proj}
\end{equation}
At a routing position $\ell_c$, the query $q_{\ell_c}$ is therefore computed
from $h_{\ell_c}$, the state produced by the preceding layers, not from the
input $E(x_{\ell_c})+r$.
The key at position $j$ is $k_j = R_j\!\left(W_K h_j\right)$ under the shared key
projection, and the content keys are the $k_j$ at content positions.
The value path and the token attention inside a selected chunk use these
backbone queries and keys.
The query that pools the chunk is $q_{\ell_c}$ with the residual of
Section~\ref{sec:topk-selection} added.

\paragraph{Chunk summary.}
Let $\mathcal{C}_c$ be the content positions of chunk $c$.
Write $\hat q_{\ell_c}$ for the query at the routing position after the
residual is added.
It attends over the content keys and pools them into one key for the chunk,
\begin{equation}
  \alpha_{c,j}
  = \operatorname{softmax}_{j\in\mathcal{C}_c}
    \!\left(\hat q_{\ell_c}^{\top} k_j / \sqrt{d}\right),
  \qquad
  s_c = \sum_{j\in\mathcal{C}_c} \alpha_{c,j}\, k_j.
  \label{eq:summary}
\end{equation}
Here $d$ is the per-head dimension of the queries and keys.
The routing position's own key is excluded from this softmax.
The summary $s_c$ serves as the key of the routing mask: it is a weighted
sum of the content keys, not the projection
$k_{\ell_c}=R_{\ell_c}(W_K h_{\ell_c})$.
Tokens whose keys have a larger inner product with $\hat q_{\ell_c}$
contribute more to $s_c$, so the
summary can concentrate on a few positions inside the chunk.
The role offset changes $\hat q_{\ell_c}$ through the hidden state at
$\ell_c$; it is not itself the summary.
A generation mask keeps $m_i=0$, and therefore keeps the mask embedding the
model was trained on.
Routing positions are omitted from the language-model targets.
The denoising loss is computed on content positions whose tokens were masked
for prediction.
The appended mask is present so that $\hat q_{\ell_c}$ can pool the chunk,
not as a token the model is asked to emit.

\subsection{Top-$K$ selection}
\label{sec:topk-selection}

\paragraph{Residual query.}
A low-rank residual is added to the backbone query at every position,
routing and content alike,
\begin{equation}
  \Delta q_i = W_{\uparrow} W_{\downarrow} h_i,
  \qquad
  \hat q_i
  = R_i\!\left(
      \operatorname{RMSNorm}(W_Q h_i + \Delta q_i)
    \right).
  \label{eq:query-residual}
\end{equation}
Here $W_Q h_i$ is the backbone query before the rotary map,
$W_{\downarrow}$ maps the model dimension down to a small rank, and
$W_{\uparrow}$ maps that rank back to the query dimension.
The sum is normalized per head with RMSNorm~\citep{zhang2019rmsnorm} before
the rotary map is applied.
At the routing position the result is $\hat q_{\ell_c}$, the query that
forms $s_c$ in Section~\ref{sec:routing-mask}.
At a content position it is $\hat q_i$, the query used in the score below.

\paragraph{Chunk score.}
Scoring reuses the summaries, so no chunk is pooled a second time.
A content position scores the summary $s_c$ of each chunk outside its window
of radius $w$ by the inner product of $\hat q_i$ with $s_c$.
The score adds an entropy term, the entropy of the pooling weights inside the
chunk,
\begin{equation}
  u_{i,c}
  = \hat q_i^{\top} s_c / \sqrt{d} + H(\alpha_c),
  \qquad
  H(\alpha_c) = -\sum_{j\in\mathcal{C}_c} \alpha_{c,j}\log\alpha_{c,j}.
  \label{eq:chunk-score}
\end{equation}
The route layer keeps the $K$ highest-scoring chunks,
\begin{equation}
  \mathcal{R}_i = \operatorname{TopK}_{c}\, u_{i,c}.
  \label{eq:topk}
\end{equation}
Chunks inside the local window, chunks without a valid routing mask, and
chunks the query is not allowed to read receive no score.
The selected set $\mathcal{R}_i$ is the remote half of what a route layer
can see.
Together with the local window of radius $w$ in
Section~\ref{sec:attn-mass}, it is the budget under which the route layer is
trained.
Each query scores $(L-2w)/S$ summaries and then reads the content tokens of
$K$ chunks, so the remote cost scales with the number of chunks outside the
window rather than with every token pair.

\paragraph{Grouped max pooling.}
Dream uses grouped-query attention (GQA)~\citep{ainsworth2023gqa}: several query heads share one key
and one value.
Those heads can prefer different chunks, but the selected tokens have to be
gathered once because the key is shared.
Each query head pools its own weights $\alpha_{c,j}$ and its own summary
$s_c$ from the keys of the shared KV head, so the summary is formed once per
query head.
Inside a GQA group, \ours takes the maximum of $u_{i,c}$ over those query
heads, chooses $\mathcal{R}_i$ from that maximum, and broadcasts the chosen
chunks to every head in the group.
The gathered values are then read once for the shared key.
Each head still receives its own weights on that shared set in
Section~\ref{sec:attn-mass}.

\subsection{Attention mass}
\label{sec:attn-mass}

\paragraph{Remote and local mass.}
Selecting $\mathcal{R}_i$ decides which chunks are visible, but not how much
attention each visible token receives.
The local branch is a bidirectional window of radius $w$ around the query,
restricted to content positions.
Its attention output is $o_i^{\mathrm{loc}}$, and the log-sum-exp of its
attention logits is $L_i^{\mathrm{loc}}$.
The selected chunk scores and this local log-sum-exp are normalized together,
\begin{equation}
  Z_i
  = \sum_{c\in\mathcal{R}_i}\exp(u_{i,c}) + \exp(L_i^{\mathrm{loc}}),
  \label{eq:partition}
\end{equation}
\begin{equation}
  w_{i,c} = \exp(u_{i,c}) / Z_i,
  \qquad
  w_i^{\mathrm{loc}} = \exp(L_i^{\mathrm{loc}}) / Z_i.
  \label{eq:mass}
\end{equation}
Thus $w_{i,c}$ is the attention mass of chunk $c$, and $w_i^{\mathrm{loc}}$
is the attention mass of the local window.
A high summary score increases the share of the attention mass that the
chunk receives, and the local window takes the remaining share.
The chunk score $u_{i,c}$ is computed with the residual query $\hat q_i$.
The local log-sum-exp $L_i^{\mathrm{loc}}$ is computed with the backbone
query $q_i$, as is the within-chunk softmax below.
Putting $u_{i,c}$ and $L_i^{\mathrm{loc}}$ in one partition assigns each
selected chunk and the local window a share of the attention mass.
This is the allocation used by the route layer, not the softmax of one
query over every visible token.

\paragraph{Within-chunk attention.}
Inside a selected chunk, token attention is a separate softmax over the
content positions,
\begin{equation}
  a_{i,j}
  = \operatorname{softmax}_{j\in\mathcal{C}_c}
    \!\left(q_i^{\top} k_j / \sqrt{d}\right),
  \qquad
  j\in\mathcal{C}_c.
  \label{eq:intra}
\end{equation}
The routing position is excluded from this softmax and from the value stream.
The summary $s_c$ is used to choose the chunk; it is not a value that can be
copied into the output.
The route-layer output combines the two stages,
\begin{equation}
  o_i
  = \sum_{c\in\mathcal{R}_i} w_{i,c}
      \sum_{j\in\mathcal{C}_c} a_{i,j}\, v_j
    + w_i^{\mathrm{loc}}\, o_i^{\mathrm{loc}}.
  \label{eq:output}
\end{equation}
Here $v_j$ is the value at position $j$ under the shared value projection.
For a token $j$ in selected chunk $c$, the product
$w_{i,c}\, a_{i,j}$ is its attention weight.
The first factor is the mass of the chunk, computed from the routing-mask
summary.
The second factor renormalizes the query--key logits inside the chunk.
Equation~\eqref{eq:output} makes the scores of
Section~\ref{sec:topk-selection} part of the forward computation.
The set $\mathcal{R}_i$ is a hard top-$K$ selection, so membership in it is
discrete and no gradient flows through the selection.
The denoising loss trains the role offset, the summary, and the low-rank
query residual through the masses $w_{i,c}$ of the chunks that were selected.

\paragraph{Non-route layers.}
On a non-route layer the same query uses only its sliding window of radius
$\rho$, with no chunk summary and no top-$K$ selection.
The route layers are where \ours spends its remote budget.
The non-route layers carry local context on both sides of each query.

\section{Experiments}
\label{sec:experiments}

\subsection{Setup}
\label{sec:setup}

Continued training starts from Dream-v0-Base-7B~\citep{ye2025dream}.
Its original context is 2K, and we extend the checkpoint to 16K with
YaRN~\citep{peng2024yarn} at a factor of 8.
We compare \ours with bidirectional NSA, DSA, and SWA.
\ours, NSA, and DSA use the same layout (three bidirectional sliding-window
layers followed by one sparse layer), the same key budget, and the same dense
initialization.
SWA applies a bidirectional sliding window in every layer.
The hyperparameters of this comparison are fixed across variants.
Appendix~\ref{sec:matched_protocol} lists the layer schedule, the window
radii, and the key budget.
Appendix~\ref{sec:dense_extrapolation} describes the dense context extension,
and Appendix~\ref{sec:continued_training} gives the continued-training
objective, the GPUs, and the settings that differ across variants.
Appendix~\ref{sec:evaluation} describes the Fast-dLLM decoder and the prompts.
At evaluation, every variant uses the same Fast-dLLM
decoder~\citep{wu2025fastdllm}.

\paragraph{Benchmarks.}
We report LongBench-v1 and three RULER tasks~\citep{bai2024longbench,hsieh2024ruler}.
For LongBench we use each task's official metric, and we report those scores
and their mean at the 16K training length.
RULER inserts the required evidence into filler text and varies the length.
Of its 13 tasks we use single-needle retrieval, multi-key multi-query, and
variable tracking, the three families supervised by the RULER term in
\eqref{eq:sync-loss}.
We do not run the official RULER release.
Training and evaluation examples both come from our own generator for these
three families, with the needle counts used in training rather than those of
the official configuration.
The filler is a document in training and the repeated RULER noise sentences at
evaluation.
At evaluation the span that the decoder fills has the same length as the gold
answer, so every model is given the answer length.
The prompts are in Appendix~\ref{sec:evaluation}.
Lengths above 16K are beyond the training length.
The other ten tasks are not reported.
An example is correct only when every gold string appears in the prediction.
We report each task and their mean at 16K, 32K, 64K, and 128K, with the
training YaRN factor held fixed and with the factor scaled to the evaluation
length.
Each number comes from a single continued-training run.
Component and routing-token ablations are in Appendix~\ref{sec:ablation}.

\begin{table*}[t]
\centering
\scriptsize
\setlength{\tabcolsep}{2pt}
\renewcommand{\arraystretch}{1.12}
\setlength{\belowcaptionskip}{8pt}
\caption{Three RULER tasks at 16K, 32K, 64K, and 128K.
The upper block is dense attention.
Full dense keeps dense attention in every layer through the same number of
continued-training steps as the other variants.
Hybrid dense uses three sliding-window layers and one dense layer.
The lower block is sparse attention.
\textit{Fixed} uses YaRN factor 8.
\textit{Scaled} sets the factor to the evaluation length divided by 2048.
MK is multi-key multi-query.
Avg.\ is the mean of S-N, MK, and VT.
In the dense block, $*$ marks the best score in each column.
In the sparse block, the best score is bold and the second best is underlined.}
\label{tab:ruler}
\resizebox{\linewidth}{!}{%
\begin{tabular}{@{}l c *{16}{c}@{}}
\toprule
&
& \multicolumn{4}{c}{\textbf{16K}}
& \multicolumn{4}{c}{\textbf{32K}}
& \multicolumn{4}{c}{\textbf{64K}}
& \multicolumn{4}{c}{\textbf{128K}} \\
\cmidrule(lr){3-6} \cmidrule(lr){7-10} \cmidrule(lr){11-14} \cmidrule(lr){15-18}
Model & YaRN
& S-N & MK & VT & Avg.
& S-N & MK & VT & Avg.
& S-N & MK & VT & Avg.
& S-N & MK & VT & Avg. \\
\midrule
\multicolumn{18}{c}{\textit{Dense}} \\
\midrule
\multirow{2}{*}{Full dense}
& \textit{Fixed}
& 100.0$^{*}$ & 100.0$^{*}$ & 91.7$^{*}$ & 97.2$^{*}$ & 100.0$^{*}$ & 96.5$^{*}$ & 75.7$^{*}$ & 90.7$^{*}$
& 94.0$^{*}$ & 68.5 & 40.7 & 67.7$^{*}$ & 85.0$^{*}$ & 90.0$^{*}$ & 34.1$^{*}$ & 69.7$^{*}$ \\
& \textit{Scaled}
& 100.0$^{*}$ & 100.0$^{*}$ & 91.7$^{*}$ & 97.2$^{*}$ & 100.0$^{*}$ & 99.5$^{*}$ & 80.0$^{*}$ & 93.2$^{*}$
& 100.0$^{*}$ & 97.5$^{*}$ & 69.3$^{*}$ & 88.9$^{*}$ & 100.0$^{*}$ & 100.0$^{*}$ & 62.3$^{*}$ & 87.4$^{*}$ \\
\cmidrule(lr){1-18}
\multirow{2}{*}{Hybrid dense}
& \textit{Fixed}
& 99.0 & 87.5 & 67.3 & 84.6 & 100.0$^{*}$ & 81.5 & 67.7 & 83.1
& 70.0 & 73.5$^{*}$ & 59.3$^{*}$ & 67.6 & 25.0 & 25.0 & 27.0 & 25.7 \\
& \textit{Scaled}
& 99.0 & 87.5 & 67.3 & 84.6 & 99.0 & 80.0 & 69.0 & 82.7
& 98.0 & 81.5 & 62.7 & 80.7 & 85.0 & 83.5 & 53.7 & 74.1 \\
\midrule
\multicolumn{18}{c}{\textit{Sparse}} \\
\midrule
\multirow{2}{*}{SWA}
& \textit{Fixed}
& 5.0 & 5.0 & 4.7 & 4.9 & 2.0 & 1.0 & 4.3 & 2.4
& 3.0 & 2.5 & 1.0 & 2.2 & 1.0 & 0.0 & 0.0 & 0.3 \\
& \textit{Scaled}
& 5.0 & 5.0 & 4.7 & 4.9 & 2.0 & 1.0 & 4.3 & 2.4
& 3.0 & 2.0 & 0.7 & 1.9 & 2.0 & 0.0 & 1.0 & 1.0 \\
\cmidrule(lr){1-18}
\multirow{2}{*}{NSA}
& \textit{Fixed}
& \underline{81.0} & \underline{32.0} & \underline{15.0} & \underline{42.7} & 84.0 & 32.0 & 10.0 & 42.0
& 79.0 & 16.0 & 9.3 & 34.8 & 79.0 & 12.0 & 5.0 & 32.0 \\
& \textit{Scaled}
& \underline{81.0} & \underline{32.0} & \underline{15.0} & \underline{42.7} & \underline{87.0} & 30.5 & 12.3 & 43.3
& 86.0 & 22.0 & 9.0 & 39.0 & 83.0 & 29.0 & \underline{9.0} & 40.3 \\
\cmidrule(lr){1-18}
\multirow{2}{*}{DSA}
& \textit{Fixed}
& 9.0 & 15.5 & 12.7 & 12.4 & 6.0 & 7.0 & 4.0 & 5.7
& 5.0 & 4.0 & 2.0 & 3.7 & 3.0 & 0.5 & 1.3 & 1.6 \\
& \textit{Scaled}
& 9.0 & 15.5 & 12.7 & 12.4 & 7.0 & 6.0 & 4.7 & 5.9
& 3.0 & 5.0 & 2.7 & 3.6 & 2.0 & 1.5 & 2.0 & 1.8 \\
\cmidrule(lr){1-18}
\rowcolor{mycolor!30}
& \textit{Fixed}
& \textbf{100.0} & \textbf{42.0} & \textbf{19.3} & \textbf{53.8} & \textbf{99.0} & \underline{35.0} & \textbf{18.0} & \underline{50.7}
& \textbf{100.0} & \underline{27.0} & \textbf{17.0} & \underline{48.0} & \textbf{99.0} & \underline{32.5} & 8.0 & \underline{46.5} \\
\rowcolor{mycolor!30}
\multirow{-2}{*}{\textbf{\ours}}
& \textit{Scaled}
& \textbf{100.0} & \textbf{42.0} & \textbf{19.3} & \textbf{53.8} & \textbf{99.0} & \textbf{47.5} & \underline{17.7} & \textbf{54.7}
& \underline{98.0} & \textbf{45.0} & \underline{14.0} & \textbf{52.3} & \underline{95.0} & \textbf{37.5} & \textbf{10.7} & \textbf{47.7} \\
\bottomrule
\end{tabular}}
\end{table*}

\subsection{Main results}
\label{sec:main_results}

\paragraph{RULER.}
Table~\ref{tab:ruler} reports the results across lengths.
Among the sparse models, \ours has the highest three-task mean at every length
and under both YaRN settings.
With the factor scaled to the evaluation length, the means are 53.8, 54.7,
52.3, and 47.7 at 16K, 32K, 64K, and 128K, against 42.7, 43.3, 39.0, and
40.3 for NSA, a margin of 11.1, 11.4, 13.3, and 7.4 points.
DSA and SWA remain far lower and reach 1.8 and 1.0 at 128K.
\ours leads NSA on single-needle retrieval, with 100.0, 99.0, 98.0, and 95.0
against 81.0, 87.0, 86.0, and 83.0, and on multi-key multi-query, with 42.0,
47.5, 45.0, and 37.5 against 32.0, 30.5, 22.0, and 29.0.
Single-needle retrieval contributes most of the margin at 16K and 128K, and
multi-key multi-query contributes most of it at 32K and 64K.
Multi-key multi-query and variable tracking remain far below full dense, 42.0
and 19.3 against 100.0 and 91.7 at 16K, because those tasks scatter the
evidence and a top-$K$ selection can miss one piece.
With the factor held at 8, the single-needle accuracy of \ours stays at 100.0
at 16K and 64K and at 99.0 at 32K and 128K, while full dense falls to 94.0 at
64K.
The fixed factor lowers multi-key accuracy, from 47.5 to 35.0 at 32K and from
45.0 to 27.0 at 64K, but the 64K mean still favors \ours, 48.0 against 34.8
for NSA.
Hybrid dense replaces the route with a dense layer and scores 84.6 at 16K,
against 97.2 for full dense.
At 128K its scaled single-needle score is 85.0, below \ours at 95.0, and the
fixed factor lowers it to 25.0.

\paragraph{LongBench.}
On LongBench the order reverses
(Table~\ref{tab:longbench_32k_main}).
The task means are 30.9 for SWA, 26.6 for NSA, 25.8 for DSA, and 25.4 for
\ours, against 44.3 for full dense.
Most of these tasks need coverage of a long document.
A route that keeps a few remote chunks drops most of that document, while
SWA applies a wide window in every layer.
The same window cannot retrieve a needle, which is why the RULER ranking
runs the other way.

\begin{table*}[t]
\centering
\scriptsize
\setlength{\tabcolsep}{2pt}
\renewcommand{\arraystretch}{1.12}
\setlength{\belowcaptionskip}{8pt}
\caption{LongBench-v1 at the 16K training length.
The upper block is dense attention.
Full dense keeps dense attention in every layer through the same number of
continued-training steps as the other variants.
Hybrid dense uses three sliding-window layers and one dense layer.
The lower block is sparse attention.
In the dense block, $*$ marks the best score in each column.
In the sparse block, the best score is bold and the second best is underlined.}
\label{tab:longbench_32k_main}
\resizebox{\linewidth}{!}{%
\begin{tabular}{@{}l *{17}{c}@{}}
\toprule
\multirow{5}{*}{Model}
& \multicolumn{3}{c}{Single-Doc.\ QA}
& \multicolumn{3}{c}{Multi-Doc.\ QA}
& \multicolumn{4}{c}{Summarization}
& \multicolumn{2}{c}{Few-shot}
& \multicolumn{2}{c}{Synthetic}
& \multicolumn{2}{c}{Code}
& \multirow{5}{*}{Avg.} \\
\cmidrule(lr){2-4}
\cmidrule(lr){5-7}
\cmidrule(lr){8-11}
\cmidrule(lr){12-13}
\cmidrule(lr){14-15}
\cmidrule(lr){16-17}
& \rotatebox{60}{NarrativeQA}
& \rotatebox{60}{Qasper}
& \rotatebox{60}{MF-en}
& \rotatebox{60}{HotpotQA}
& \rotatebox{60}{2WikiMQA}
& \rotatebox{60}{Musique}
& \rotatebox{60}{GovReport}
& \rotatebox{60}{QMSum}
& \rotatebox{60}{MultiNews}
& \rotatebox{60}{SAMSum}
& \rotatebox{60}{TREC}
& \rotatebox{60}{TriviaQA}
& \rotatebox{60}{PCount}
& \rotatebox{60}{PRe}
& \rotatebox{60}{LCC}
& \rotatebox{60}{RB-P}
& \\
\midrule
\multicolumn{18}{c}{\textit{Dense}} \\
\midrule
Full dense
& 29.1$^{*}$ & 36.0 & 53.7$^{*}$ & 60.1$^{*}$ & 47.1$^{*}$ & 38.8 & 5.1 & 22.8 & 4.7$^{*}$ & 42.9$^{*}$ & 73.5$^{*}$ & 91.2$^{*}$ & 2.0$^{*}$ & 99.5$^{*}$ & 47.9$^{*}$ & 53.8 & 44.3$^{*}$ \\
\cmidrule(lr){1-18}
Hybrid dense
& 25.4 & 37.1$^{*}$ & 48.5 & 59.3 & 40.6 & 39.2$^{*}$ & 8.7$^{*}$ & 24.9$^{*}$ & 1.5 & 35.7 & 72.0 & 88.7 & 2.0$^{*}$ & 96.3 & 46.2 & 55.7$^{*}$ & 42.6 \\
\midrule
\multicolumn{18}{c}{\textit{Sparse}} \\
\midrule
SWA
& \underline{18.0} & \textbf{20.8} & \underline{26.4} & \textbf{28.3} & \textbf{28.1} & \underline{12.8} & \textbf{19.8} & \textbf{20.7} & 13.7 & \textbf{41.9} & \textbf{64.0} & \textbf{86.2} & \underline{1.7} & \textbf{14.5} & \textbf{45.8} & \textbf{50.8} & \textbf{30.9} \\
\cmidrule(lr){1-18}
NSA
& 13.1 & 16.2 & \textbf{28.3} & 22.6 & 20.6 & 10.8 & \underline{16.2} & \underline{20.6} & \underline{17.0} & \underline{38.9} & 60.0 & \underline{79.5} & 0.3 & 0.5 & \underline{38.6} & 41.9 & \underline{26.6} \\
\cmidrule(lr){1-18}
DSA
& \textbf{18.1} & \underline{17.3} & \textbf{28.3} & 20.4 & 22.7 & \textbf{13.3} & 15.2 & 16.5 & 14.0 & 34.2 & 61.0 & 74.8 & \textbf{2.1} & 4.4 & 33.4 & 36.9 & 25.8 \\
\cmidrule(lr){1-18}
\rowcolor{mycolor!30}
\textbf{\ours}
& 10.4 & 14.8 & 17.7 & \underline{24.3} & \underline{26.3} & 10.1 & 13.4 & 18.4 & \textbf{17.2} & 35.0 & \underline{62.0} & 70.6 & 0.8 & \underline{5.6} & 37.6 & \underline{42.6} & 25.4 \\
\bottomrule
\end{tabular}}
\end{table*}

\subsection{Routing-mask probes}
\label{sec:mask-probe}

\begin{figure*}[t]
\centering
\includegraphics[width=0.6\textwidth]{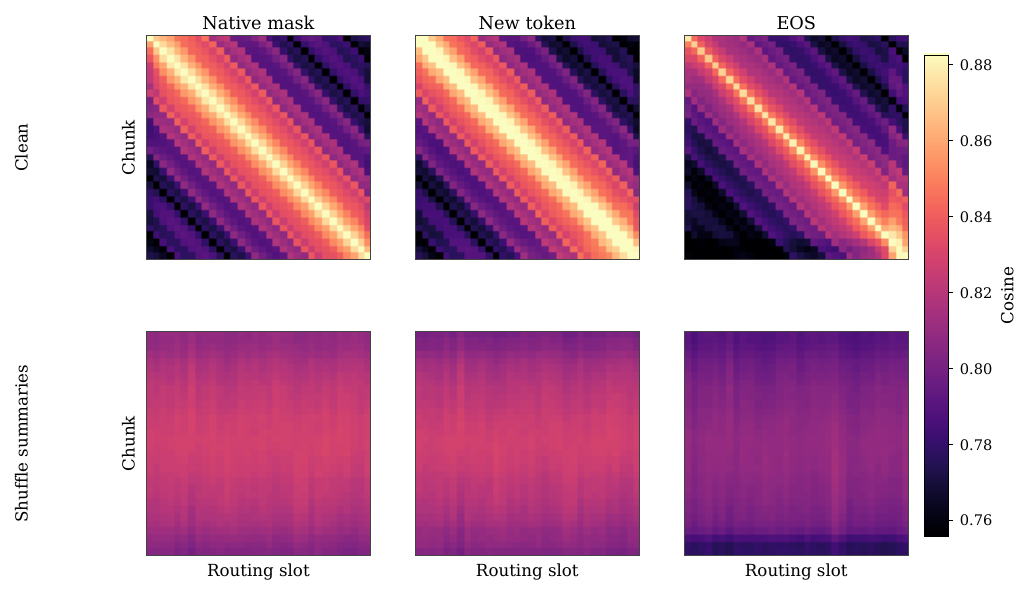}
\caption{Cosine between the mean content key of each chunk and the summary
key of each routing slot on the 16K single-needle probe, averaged over route
layers and 100 examples.
Each panel is binned to $32\times 32$.
On the clean probe the diagonal pairs a slot with its own chunk.
\emph{Shuffle summaries} permutes the summaries across chunks.}
\label{fig:mask-heat}
\end{figure*}

\begin{figure*}[t]
\centering
\includegraphics[width=0.6\textwidth]{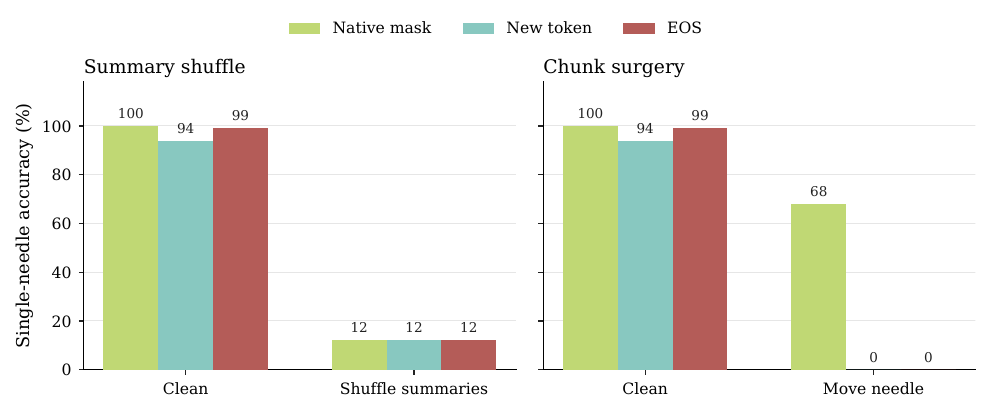}
\caption{16K single-needle accuracy for the routing mask, a new learned token,
and EOS.
\emph{Shuffle summaries} permutes the summaries across chunks and leaves the
text in place.
\emph{Move needle} places the intact needle in another chunk.
Each bar is 100 examples.}
\label{fig:mask-probe}
\end{figure*}

Figure~\ref{fig:mask-probe} tests the routing mask on 16K single-needle
retrieval (100 examples) with the same Fast-dLLM decoder.
In addition to the native mask, we evaluate two routing slots trained the same
way: the end-of-sequence (EOS) token and a new learned token.
In the first probe we permute the summaries across chunks and leave the text
in place.
Accuracy falls from 100.0, 94.0, and 99.0 to 12.0 for the mask, the new token,
and EOS.
Figure~\ref{fig:mask-heat} shows why.
On the clean probe the cosine between the mean content key of a chunk and the
summary key of a routing slot concentrates on the diagonal, so each summary
matches its own chunk.
The new token and EOS show the same diagonal.
Permuting the summaries removes it.
The needle is still in the text, and breaking the binding between each summary
and its chunk is enough to lose it.

\textbf{Finding.}
The routing mask stores a chunk summary that the model reads.

In the second probe we move the intact needle into another chunk.
The routing slots stay where they are, and only the needle span is exchanged
with the tokens at the destination.
Accuracy falls to 68.0 for the native mask and to 0.0 for the new token and
for EOS.
The mask summary is formed again from the chunk that now holds the needle, and
the route layer still selects that chunk.
A new token and EOS match the mask when the needle is not moved, and lose the
needle once it moves.

\textbf{Finding.}
After the needle moves to another chunk, the routing mask still finds it in
68 of 100 examples; a new token and EOS find it in none.

\subsection{Latency}
\label{sec:latency}

\begin{figure*}[t]
\centering
\includegraphics[width=0.6\textwidth]{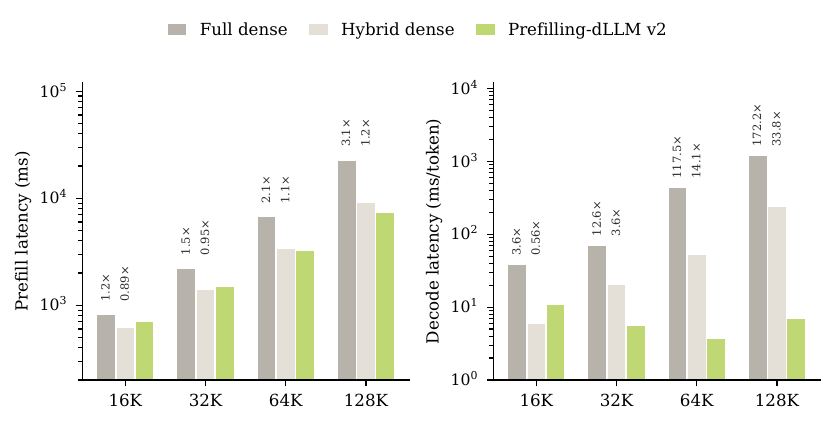}
\caption{Prefill time and decode time per generated token under the same
Fast-dLLM decoder.
Both vertical axes are logarithmic.
The number above each full-dense and hybrid-dense bar is the ratio of that
latency to the latency of \ours.
The YaRN factor is the evaluation length divided by 2048.}
\label{fig:latency}
\end{figure*}

Figure~\ref{fig:latency} reports Fast-dLLM prefill and decode latency for full
dense, hybrid dense, and \ours.
Each bar is the median of ten trials on one synthetic sequence with 128 masked
answer tokens, and the YaRN factor is the evaluation length divided by 2048.
NSA, DSA, and SWA are not timed.
At 128K, prefill is 22338\,ms for full dense, 9034\,ms for hybrid dense, and
7280\,ms for \ours, so \ours is $3.1\times$ faster than full dense; the gain is
modest because the prefill pass still reads every token.
The gap opens in per-token decoding.
Full dense rises from 37.66\,ms at 16K to 1190\,ms at 128K.
Hybrid dense is faster than \ours at 16K, 5.90\,ms against 10.55\,ms, then
reaches 233.4\,ms at 128K, $5.1\times$ lower than full dense.
\ours stays between 3.69\,ms and 10.55\,ms per token and is 6.91\,ms at 128K,
$172\times$ lower than full dense, because the remote budget is a fixed number
of chunks.

\textbf{Finding.}
At long context the prefill gain is much smaller than the decode gain:
$3.1\times$ against $172\times$ at 128K.

\section{Conclusion}

\ours continues training a diffusion language model so that it routes with its native mask.
A mask at the end of each chunk summarizes that chunk, and every fourth layer
keeps a local window plus the top-$K$ chunks scored from that summary.
Under a matched budget and the same Fast-dLLM decoder, \ours leads NSA, DSA,
and SWA on all three RULER tasks from 16K to 128K when the YaRN factor scales
with length.
At 128K, decoding is $172\times$ faster than with full dense attention,
because the remote budget is a fixed number of chunks.

Those three tasks are the task families in the continued-training loss, and
lengths beyond 16K were not seen in training.
Multi-key retrieval and variable tracking still trail full dense, because
missing one chunk is enough to get an example wrong when its evidence is
scattered.
On LongBench, where most of a document is useful, a wide window outperforms a few
selected chunks.
A fixed routing budget is enough to retrieve a localized needle, and it keeps
decoding cheap as the context grows.

\bibliographystyle{iclr2027_conference}
\bibliography{refs}

@inproceedings{austin2021d3pm,
  title     = {Structured Denoising Diffusion Models in Discrete State-Spaces},
  author    = {Austin, Jacob and Johnson, Daniel D. and Ho, Jonathan and Tarlow, Daniel and van den Berg, Rianne},
  booktitle = {Advances in Neural Information Processing Systems},
  year      = {2021}
}

@inproceedings{sahoo2024mdlm,
  title     = {Simple and Effective Masked Diffusion Language Models},
  author    = {Sahoo, Subham Sekhar and Arriola, Marianne and Schiff, Yair and Gokaslan, Aaron and Marroquin, Edgar and Chiu, Justin T. and Rush, Alexander and Kuleshov, Volodymyr},
  booktitle = {Advances in Neural Information Processing Systems},
  year      = {2024}
}

@article{nie2025llada,
  title   = {{LLaDA}: Large Language Diffusion Models},
  author  = {Nie, Shen and Zhu, Fengqi and You, Zebin and Zhang, Xiaolu and Ou, Jingyang and Hu, Jun and Zhou, Jun and Lin, Yankai and Wen, Ji-Rong and Li, Chongxuan},
  journal = {arXiv preprint arXiv:2502.09992},
  year    = {2025}
}

@article{ye2025dream,
  title   = {Dream {7B}: Diffusion Large Language Models},
  author  = {Ye, Jiacheng and Xie, Zhihui and Zheng, Lin and Gao, Jiahui and Wu, Zirui and Jiang, Xin and Li, Zhenguo and Kong, Lingpeng},
  journal = {arXiv preprint arXiv:2508.15487},
  year    = {2025}
}

@inproceedings{li2022diffusionlm,
  title     = {Diffusion-{LM} Improves Controllable Text Generation},
  author    = {Li, Xiang Lisa and Thickstun, John and Gulrajani, Ishaan and Liang, Percy and Hashimoto, Tatsunori B.},
  booktitle = {Advances in Neural Information Processing Systems},
  year      = {2022}
}

@inproceedings{gong2023diffseq,
  title     = {{DiffuSeq}: Sequence to Sequence Text Generation with Diffusion Models},
  author    = {Gong, Shansan and Li, Mukai and Feng, Jiangtao and Wu, Zhiyong and Kong, Lingpeng},
  booktitle = {International Conference on Learning Representations},
  year      = {2023}
}

@inproceedings{he2023diffusionbert,
  title     = {{DiffusionBERT}: Improving Generative Masked Language Models with Diffusion Models},
  author    = {He, Zhengfu and Sun, Tianxiang and Wang, Kuanning and Huang, Xuanjing and Qiu, Xipeng},
  booktitle = {Proceedings of the 61st Annual Meeting of the Association for Computational Linguistics},
  year      = {2023}
}

@inproceedings{gong2025adapt,
  title     = {Scaling Diffusion Language Models via Adaptation from Autoregressive Models},
  author    = {Gong, Shansan and Agarwal, Shivam and Zhang, Yizhe and Ye, Jiacheng and Zheng, Lin and Li, Mukai and An, Chenxin and Zhao, Peilin and Bi, Wei and Han, Jiawei and Peng, Hao and Kong, Lingpeng},
  booktitle = {International Conference on Learning Representations},
  year      = {2025}
}

@article{zhu2025llada15,
  title   = {{LLaDA} 1.5: Variance-Reduced Preference Optimization for Large Language Diffusion Models},
  author  = {Zhu, Fengqi and Wang, Rongzhen and Nie, Shen and Zhang, Xiaolu and Wu, Chunwei and Hu, Jun and Zhou, Jun and Chen, Jianfei and Lin, Yankai and Wen, Ji-Rong and Li, Chongxuan},
  journal = {arXiv preprint arXiv:2505.19223},
  year    = {2025}
}

@article{you2025lladav,
  title   = {{LLaDA-V}: Large Language Diffusion Models with Visual Instruction Tuning},
  author  = {You, Zebin and Nie, Shen and Zhang, Xiaolu and Hu, Jun and Zhou, Jun and Lu, Zhiwu and Wen, Ji-Rong and Li, Chongxuan},
  journal = {arXiv preprint arXiv:2505.16933},
  year    = {2025}
}

@article{liu2025longllada,
  title   = {{LongLLaDA}: Unlocking Long Context Capabilities in Diffusion {LLMs}},
  author  = {Liu, Xiaoran and Song, Yuerong and Liu, Zhigeng and Huang, Zengfeng and Guo, Qipeng and He, Ziwei and Qiu, Xipeng},
  journal = {arXiv preprint arXiv:2506.14429},
  year    = {2025}
}

@article{arriola2025bd3lm,
  title   = {Block Diffusion: Interpolating Between Autoregressive and Diffusion Language Models},
  author  = {Arriola, Marianne and Gokaslan, Aaron and Chiu, Justin T. and Yang, Zhihan and Qi, Zhixuan and Han, Jiaqi and Sahoo, Subham Sekhar and Kuleshov, Volodymyr},
  journal = {arXiv preprint arXiv:2503.09573},
  year    = {2025}
}

@article{wu2025fastdllmv2,
  title   = {Fast-{dLLM} v2: Efficient Block-Diffusion {LLM}},
  author  = {Wu, Chengyue and Zhang, Hao and Xue, Shuchen and Diao, Shizhe and Fu, Yonggan and Liu, Zhijian and Molchanov, Pavlo and Luo, Ping and Han, Song and Xie, Enze},
  journal = {arXiv preprint arXiv:2509.26328},
  year    = {2025}
}

@article{wu2025fastdllm,
  title   = {Fast-{dLLM}: Training-free Acceleration of Diffusion {LLM} by Enabling {KV} Cache and Parallel Decoding},
  author  = {Wu, Chengyue and Zhang, Hao and Xue, Shuchen and Liu, Zhijian and Diao, Shizhe and Zhu, Ligeng and Luo, Ping and Han, Song and Xie, Enze},
  journal = {arXiv preprint arXiv:2505.22618},
  year    = {2025}
}

@article{xiong2026prefilling,
  title   = {Prefilling-{dLLM}: Predictive Prefilling for Long-Context Inference in Diffusion Language Models},
  author  = {Xiong, Jing and Han, Qi and Gong, Shansan and Hsieh, Yunta and Wu, Chengyue and Tao, Chaofan and Zhao, Chenyang and Wong, Ngai},
  journal = {arXiv preprint arXiv:2606.10537},
  year    = {2026}
}

@inproceedings{xiong2025parallelcomp,
  title     = {{ParallelComp}: Parallel Long-Context Compressor for Length Extrapolation},
  author    = {Xiong, Jing and Shen, Jianghan and Zheng, Chuanyang and Wan, Zhongwei and Zhao, Chenyang and Yang, Chiwun and Ye, Fanghua and Yang, Hongxia and Kong, Lingpeng and Wong, Ngai},
  booktitle = {Proceedings of the 42nd International Conference on Machine Learning},
  year      = {2025}
}

@inproceedings{xiong2025uncomp,
  title     = {{UNComp}: Can Matrix Entropy Uncover Sparsity? --- {A} Compressor Design from an Uncertainty-Aware Perspective},
  author    = {Xiong, Jing and Shen, Jianghan and Ye, Fanghua and Tao, Chaofan and Wan, Zhongwei and Lu, Jianqiao and Wu, Xun and Zheng, Chuanyang and Guo, Zhijiang and Yang, Min and Kong, Lingpeng and Wong, Ngai},
  booktitle = {Proceedings of the 2025 Conference on Empirical Methods in Natural Language Processing},
  year      = {2025}
}

@inproceedings{yuan2025nsa,
  title     = {Native Sparse Attention: Hardware-Aligned and Natively Trainable Sparse Attention},
  author    = {Yuan, Jingyang and Gao, Huazuo and Dai, Damai and Luo, Junyu and Zhao, Liang and Zhang, Zhengyan and Xie, Zhenda and Wei, Y. X. and Wang, Lean and Xiao, Zhiping and Wang, Yuqing and Ruan, Chong and Zhang, Ming and Liang, Wenfeng and Zeng, Wangding},
  booktitle = {Proceedings of the 63rd Annual Meeting of the Association for Computational Linguistics (Volume 1: Long Papers)},
  year      = {2025}
}

@article{deepseek2025v32,
  title   = {{DeepSeek-V3.2}: Pushing the Frontier of Open Large Language Models},
  author  = {{DeepSeek-AI}},
  journal = {arXiv preprint arXiv:2512.02556},
  year    = {2025}
}

@article{hu2026hils,
  title   = {Hierarchical Sparse Attention Done Right: Toward Infinite Context Modeling},
  author  = {Hu, Xiang and Wei, Xinyu and Gu, Hao and Zhang, Minshen and Liang, Tian and Li, Huayang and Zhu, Lei and Wang, Yan and Han, Sirui and Bai, Yushi and Tu, Kewei and Mi, Haitao and Liang, Leo},
  journal = {arXiv preprint arXiv:2607.02980},
  year    = {2026}
}

@inproceedings{hsieh2024ruler,
  title     = {{RULER}: What's the Real Context Size of Your Long-Context Language Models?},
  author    = {Hsieh, Cheng-Ping and Sun, Simeng and Kriman, Samuel and Acharya, Shantanu and Rekesh, Dima and Jia, Fei and Zhang, Yang and Ginsburg, Boris},
  booktitle = {Conference on Language Modeling},
  year      = {2024}
}

@article{wang2025sparsed,
  title   = {{SparseD}: Sparse Attention for Diffusion Language Models},
  author  = {Wang, Zeqing and Fang, Gongfan and Ma, Xinyin and Yang, Xingyi and Wang, Xinchao},
  journal = {arXiv preprint arXiv:2509.24014},
  year    = {2025}
}

@article{song2025sparsedllm,
  title   = {Sparse-{dLLM}: Accelerating Diffusion {LLMs} with Dynamic Cache Eviction},
  author  = {Song, Yuerong and Liu, Xiaoran and Li, Ruixiao and Liu, Zhigeng and Huang, Zengfeng and Guo, Qipeng and He, Ziwei and Qiu, Xipeng},
  journal = {arXiv preprint arXiv:2508.02558},
  year    = {2025}
}

@article{peng2024yarn,
  title   = {{YaRN}: Efficient Context Window Extension of Large Language Models},
  author  = {Peng, Bowen and Quesnelle, Jeffrey and Fan, Honglu and Shippole, Enrico},
  journal = {arXiv preprint arXiv:2309.00071},
  year    = {2023}
}

@article{bai2024longbench,
  title   = {{LongBench}: A Bilingual, Multitask Benchmark for Long Context Understanding},
  author  = {Bai, Yushi and Lv, Xin and Zhang, Jiajie and Lyu, Hongchang and Tang, Jiankai and Huang, Zhidian and Du, Zhengxiao and Liu, Xiao and Zeng, Aohan and Hou, Lei and Dong, Yuxiao and Tang, Jie and Li, Juanzi},
  journal = {arXiv preprint arXiv:2308.14508},
  year    = {2023}
}

@inproceedings{xiao2024streamingllm,
  title     = {Efficient Streaming Language Models with Attention Sinks},
  author    = {Xiao, Guangxuan and Tian, Yuandong and Chen, Beidi and Han, Song and Lewis, Mike},
  booktitle = {International Conference on Learning Representations},
  year      = {2024}
}

@article{jiang2023mistral,
  title   = {Mistral 7{B}},
  author  = {Jiang, Albert Q. and Sablayrolles, Alexandre and Mensch, Arthur and Bamford, Chris and Chaplot, Devendra Singh and Casas, Diego de las and Bressand, Florian and Lengyel, Gianna and Lample, Guillaume and Saulnier, Lucile and Lavaud, L{\'e}lio Renard and Lachaux, Marie-Anne and Stock, Pierre and Scao, Teven Le and Lavril, Thibaut and Wang, Thomas and Lacroix, Timoth{\'e}e and Sayed, William El},
  journal = {arXiv preprint arXiv:2310.06825},
  year    = {2023}
}

@article{child2019sparsetransformer,
  title   = {Generating Long Sequences with Sparse Transformers},
  author  = {Child, Rewon and Gray, Scott and Radford, Alec and Sutskever, Ilya},
  journal = {arXiv preprint arXiv:1904.10509},
  year    = {2019}
}

@article{beltagy2020longformer,
  title   = {Longformer: The Long-Document Transformer},
  author  = {Beltagy, Iz and Peters, Matthew E. and Cohan, Arman},
  journal = {arXiv preprint arXiv:2004.05150},
  year    = {2020}
}

@inproceedings{zaheer2020bigbird,
  title     = {{BigBird}: Transformers for Longer Sequences},
  author    = {Zaheer, Manzil and Guruganesh, Guru and Dubey, Kumar Avinava and Ainslie, Joshua and Alberti, Chris and Ontanon, Santiago and Pham, Philip and Ravula, Anirudh and Wang, Qifan and Yang, Li and Ahmed, Amr},
  booktitle = {Advances in Neural Information Processing Systems},
  year      = {2020}
}

@article{lu2025moba,
  title   = {{MoBA}: Mixture of Block Attention for Long-Context {LLMs}},
  author  = {Lu, Enzhe and Jiang, Zhejun and Liu, Jingyuan and Du, Yulun and Jiang, Tao and Hong, Chao and Liu, Shaowei and He, Weiran and Yuan, Enming and Wang, Yuzhi and Huang, Zhiqi and Yuan, Huan and Xu, Suting and Xu, Xinran and Lai, Guokun and Chen, Yanru and Zheng, Huabin and Yan, Junjie and Su, Jianlin and Wu, Yuxin and Zhang, Yutao and Yang, Zhilin and Zhou, Xinyu and Zhang, Mingxing and Qiu, Jiezhong},
  journal = {arXiv preprint arXiv:2502.13189},
  year    = {2025}
}

@inproceedings{zhang2023h2o,
  title     = {{H2O}: Heavy-Hitter Oracle for Efficient Generative Inference of Large Language Models},
  author    = {Zhang, Zhenyu and Sheng, Ying and Zhou, Tianyi and Chen, Tianlong and Zheng, Lianmin and Cai, Ruisi and Song, Zhao and Tian, Yuandong and R{\'e}, Christopher and Barrett, Clark and Wang, Zhangyang and Chen, Beidi},
  booktitle = {Proceedings of the 40th International Conference on Machine Learning},
  year      = {2023}
}

@inproceedings{jiang2024minference,
  title     = {{MInference} 1.0: Accelerating Pre-filling for Long-Context {LLMs} via Dynamic Sparse Attention},
  author    = {Jiang, Huiqiang and Li, Yucheng and Zhang, Chengruidong and Wu, Qianhui and Luo, Xufang and Ahn, Surin and Han, Zhenhua and Abdi, Amir H. and Li, Dongsheng and Lin, Chin-Yew and Yang, Yuqing and Qiu, Lili},
  booktitle = {Advances in Neural Information Processing Systems},
  year      = {2024}
}

@inproceedings{ainsworth2023gqa,
  title     = {{GQA}: Training Generalized Multi-Query Transformer Models from Multi-Head Checkpoints},
  author    = {Ainslie, Joshua and Lee-Thorp, James and de Jong, Michiel and Zemlyanskiy, Yury and Lebr{\'o}n, Federico and Sanghai, Sumit},
  booktitle = {Proceedings of the 2023 Conference on Empirical Methods in Natural Language Processing},
  year      = {2023}
}

@inproceedings{zhang2019rmsnorm,
  title     = {Root Mean Square Layer Normalization},
  author    = {Zhang, Biao and Sennrich, Rico},
  booktitle = {Advances in Neural Information Processing Systems},
  year      = {2019}
}

@inproceedings{hu2022lora,
  title     = {{LoRA}: Low-Rank Adaptation of Large Language Models},
  author    = {Hu, Edward J. and Shen, Yelong and Wallis, Phillip and Allen-Zhu, Zeyuan and Li, Yuanzhi and Wang, Shean and Wang, Lu and Chen, Weizhu},
  booktitle = {International Conference on Learning Representations},
  year      = {2022}
}

@article{olmo2025olmo3,
  title   = {Olmo 3},
  author  = {{Team Olmo} and Ettinger, Allyson and Bertsch, Amanda and Kuehl, Bailey and Graham, David and Heineman, David and Groeneveld, Dirk and others},
  journal = {arXiv preprint arXiv:2512.13961},
  year    = {2025}
}
\newpage
\appendix
\section{Experimental setup}
\label{sec:experimental_setup}

This appendix lists the settings used in Section~\ref{sec:experiments}.

\subsection{Basic setup}
\label{sec:matched_protocol}

These experiments use NVIDIA H20 GPUs.
Training and evaluation run in PyTorch 2.9.1 built with CUDA 12.8.
The model is Dream-v0-Base-7B, with 28 transformer layers.
Its original context is 2048 tokens, and YaRN with factor 8 extends
the checkpoint to 16K.

Table~\ref{tab:contracts} lists the attention configuration of each variant.
Layers $\{3,7,11,15,19,23,27\}$, every fourth layer, are route layers for
\ours, NSA, and DSA, and dense layers for hybrid dense.
Each of the other 21 layers applies its own bidirectional sliding window of
radius $\rho=1280$ around every query, about 2561 positions.
Full dense keeps dense attention in every layer, and SWA uses radius 1280
in every layer.
On a \ours route layer the nominal budget is about 2560 keys: a
bidirectional window of radius $w=256$ together with $K=32$ selected chunks
of $S-1=63$ content tokens, before overlap between the window and the
selected chunks.
Overlap at chunk boundaries can make the number of distinct keys slightly
smaller.

\begin{table*}[t]
\centering
\small
\setlength{\tabcolsep}{4pt}
\caption{Layer configurations.
Each variant occupies two rows.
The first row is every fourth layer, and the second row is the other 21
layers.
Local window is a bidirectional radius.
Remote budget is the set of keys gathered beyond that window.
Total budget is the number of key positions on that row.
The sparse rows share a budget of about 2560.
A dense layer reads every key.}
\label{tab:contracts}
\begin{tabular}{@{}llllc@{}}
\toprule
Variant & Layers & Local window & Remote budget & Total budget \\
\midrule
\multirow{2}{*}{\ours}
& every 4th & radius 256 & 32 chunks of 64 & $\approx 2560$ \\
& other 21 & radius 1280 & -- & $\approx 2560$ \\
\midrule
\multirow{2}{*}{NSA}
& every 4th & radius 128 & $32\times 64$ selected, one token per 64 & $\approx 2560$ \\
& other 21 & radius 1280 & -- & $\approx 2560$ \\
\midrule
\multirow{2}{*}{DSA}
& every 4th & -- & token top-$k$ 2560 & 2560 \\
& other 21 & radius 1280 & -- & $\approx 2560$ \\
\midrule
\multirow{2}{*}{SWA}
& every 4th & radius 1280 & -- & $\approx 2560$ \\
& other 21 & radius 1280 & -- & $\approx 2560$ \\
\midrule
\multirow{2}{*}{Hybrid dense}
& every 4th & all keys & -- & all keys \\
& other 21 & radius 1280 & -- & $\approx 2560$ \\
\midrule
\multirow{2}{*}{Full dense}
& every 4th & all keys & -- & all keys \\
& other 21 & all keys & -- & all keys \\
\bottomrule
\end{tabular}
\end{table*}

\subsection{Dense context extension}
\label{sec:dense_extrapolation}

This stage continues training Dream-v0-Base-7B with dense attention in every
layer.
YaRN with factor 8 extends the original context of 2048 to 16384
positions.
The run uses two GPUs.

The language-model data are document windows from Dolma 3
Longmino~\citep{olmo2025olmo3}.\footnote{\url{https://huggingface.co/datasets/allenai/dolma3_longmino_mix-50B-1025}}
Each window is one contiguous span of a single document, 16128 tokens long.
This is the amount of text that fills a 16384-position window when one
position in every block of 64 is reserved.
The corpus has 3200 windows.
We hold out 128 windows for validation, and the remaining windows form the
training set.
Documents shorter than one window are dropped.

A training example is a language-model window with probability 0.95 and a
RULER window with probability 0.05.
On a language-model window, the mask rate $t$ is drawn uniformly from
$[0.2, 0.8]$.
One view masks eligible tokens at that rate, and the second view masks the
remaining eligible tokens, so the two views supervise each of those tokens
once.
An eligible token is a token whose preceding position is also inside the
document and is not an end-of-sequence token.
On a RULER window, the same document is the filler.
The task is drawn uniformly from single-needle retrieval, multi-key
multi-query, and variable tracking, and the loss supervises the answer
tokens.
The objective is the mean cross-entropy over supervised tokens,
\begin{equation}
  \mathcal{L}
  = \operatorname{mean}[\mathrm{CE}].
\end{equation}

The base weights stay frozen.
Low-rank adapters~\citep{hu2022lora} of rank 16, scale 32, and dropout
0.05 train the query, key, value, and output projections.
Optimization runs for 500 steps with a cosine schedule at learning rate
$2\times 10^{-4}$,
with 25 warmup steps, zero weight decay, and gradient clipping at 1.0.
Each GPU holds one sequence per step.
Table~\ref{tab:dense_extrapolation} collects these settings.

\begin{table}[t]
\centering
\small
\setlength{\tabcolsep}{6pt}
\caption{Dense context extension.
The run continues Dream-v0-Base-7B under dense attention.
The base weights stay frozen, and the listed low-rank adapters are the
trainable parameters.}
\label{tab:dense_extrapolation}
\begin{tabular}{@{}ll@{}}
\toprule
Hyperparameter & Value \\
\midrule
GPUs & 2 \\
Steps & 500 \\
Text tokens per sequence & 16128 \\
YaRN factor & 8 \\
Optimizer & AdamW \\
Learning rate & $2\times 10^{-4}$ \\
Schedule & cosine \\
Warmup steps & 25 \\
Weight decay & 0 \\
Gradient clipping & 1.0 \\
Micro-batch size & 1 \\
Gradient accumulation & 1 \\
LoRA rank & 16 \\
LoRA scale & 32 \\
LoRA dropout & 0.05 \\
LoRA modules & query, key, value, output \\
Mask rate & $[0.2,\,0.8]$ \\
RULER probability & 0.05 \\
Seed & 7 \\
\bottomrule
\end{tabular}
\end{table}

\subsection{Continued training}
\label{sec:continued_training}

The six variants in the main comparison continue from the step-500
checkpoint of Section~\ref{sec:dense_extrapolation}.
They use the same Dolma 3 Longmino windows.
Every step contains both a language-model view and a RULER view.
The language-model view keeps the two complementary masks of that section,
with the mask rate drawn uniformly from $[0.2, 0.8]$.
The RULER view draws one task uniformly from single-needle retrieval,
multi-key multi-query, and variable tracking, and supervises the answer
tokens.
The two views are averaged as token means,
\begin{equation}
  \mathcal{L}
  = \tfrac{1}{2}\,\operatorname{mean}\!\left[\mathrm{CE}_{\mathrm{Dolma}}\right]
    + \tfrac{1}{2}\,\operatorname{mean}\!\left[\mathrm{CE}_{\mathrm{RULER}}\right].
  \label{eq:sync-loss}
\end{equation}
Each run uses four GPUs, one sequence per GPU, for 1000 steps with a cosine
schedule at learning rate $10^{-4}$ and 50 warmup steps.
The base weights stay frozen, and low-rank adapters of rank 16, scale
32, and dropout 0.05 train the query, key, value, and output projections.
Table~\ref{tab:continued_training} records the parameters that differ.

\begin{table*}[t]
\centering
\small
\setlength{\tabcolsep}{6pt}
\caption{Continued training.
Each variant is initialized from the dense step-500 checkpoint and trains
for 1000 steps with a cosine schedule on four GPUs, at learning rate
$10^{-4}$ with 50 warmup steps.
The base weights stay frozen.
Low-rank adapters of rank 16, scale 32, and dropout 0.05 train the
query, key, value, and output projections.
The rows below are the parameters that differ across variants.}
\label{tab:continued_training}
\begin{tabular}{@{}lll@{}}
\toprule
Variant & Hyperparameter & Value \\
\midrule
\multirow{6}{*}{NSA}
& Route layers & every 4th \\
& Compressed branch & block 64, stride 64 \\
& Selected branch & 32 blocks of 64 \\
& Local branch & radius 128 \\
& Gate & three sigmoids per head \\
& Other layers & SWA radius 1280 \\
\midrule
\multirow{8}{*}{DSA}
& Route layers & every 4th \\
& Warmup & 100 steps, dense attention \\
& Warmup objective & indexer KL on every query \\
& Indexer learning rate & $10^{-3}$ \\
& Indexer & 8 heads, dimension 128 \\
& Sparse phase & 900 steps, token top-$k$ 2560 \\
& Sparse KL & weight 1, 64 queries, selected tokens \\
& Other layers & SWA radius 1280 \\
\midrule
\multirow{2}{*}{SWA}
& Layers & all 28 \\
& Window & radius 1280 \\
\midrule
\multirow{7}{*}{\ours}
& Route layers & every 4th \\
& Local window & radius 256 \\
& Remote budget & 32 chunks of 64 \\
& Selection & hard top-$K$ \\
& Role offset & one learned vector, added at the input \\
& Query residual & rank 64 \\
& Other layers & SWA radius 1280 \\
\midrule
\multirow{2}{*}{Hybrid dense}
& Every 4th layer & dense, all keys \\
& Other layers & SWA radius 1280 \\
\midrule
\multirow{2}{*}{Full dense}
& Layers & all 28 \\
& Attention & dense, all keys \\
\bottomrule
\end{tabular}
\end{table*}

\paragraph{NSA.}
Every fourth layer mixes three branches.
The compressed branch maps each non-overlapping block of 64 tokens to one
key and one value.
The selected branch gathers 32 of those blocks, 64 tokens each, using the
attention weights of the compressed branch.
The local branch is a bidirectional window of radius 128.
A learned gate gives each head three independent sigmoids and mixes the three
outputs with those weights.
The gate and the compression parameters train together with the low-rank
adapters.
The other 21 layers use a bidirectional window of radius 1280.

\paragraph{DSA.}
Every fourth layer retrieves individual tokens, and training has two phases.
During the first 100 steps those layers stay dense and only the indexer is
updated, at learning rate $10^{-3}$.
The objective in this phase is a KL divergence that matches the indexer to the
dense attention over the indexer's top 2560 tokens, computed for every
query.
The indexer has 8 heads of dimension 128, and all of its weights are
trained.
During the remaining 900 steps those layers attend only to the top 2560
tokens chosen by the indexer.
The language-model loss is restored, and the KL term is kept at weight 1; it
is now computed on 64 queries per sequence, still over the selected tokens.
In this phase the indexer and the low-rank adapters share a cosine schedule at
learning rate $10^{-4}$ with 50 warmup steps.
The other 21 layers use a bidirectional window of radius 1280.

\paragraph{SWA.}
Every layer uses a bidirectional window of radius 1280.
The trainable parameters are the low-rank adapters.

\paragraph{\ours.}
Every fourth layer is a route layer, and the other 21 layers use a
bidirectional window of radius 1280.
Each chunk appends the native mask, and a learned role offset is added to
that mask once, at the model input.
A rank-64 residual is added to the backbone query before the rotary map.
Chunk selection is hard top-$K$: a local window of radius 256 together with
32 chunks of 64 positions.
The denoising loss trains the role offset and the residual together with the
low-rank adapters.
Routing positions are left out of the language-model targets.

\paragraph{Hybrid dense.}
Every fourth layer reads every key.
The other 21 layers use a bidirectional window of radius 1280.
The trainable parameters are the low-rank adapters.

\paragraph{Full dense.}
Every layer reads every key.
The trainable parameters are the low-rank adapters.

\subsection{Evaluation}
\label{sec:evaluation}

Every variant uses the same Fast-dLLM decoder~\citep{wu2025fastdllm}.
Key and value states are cached and reused across denoising steps.
A route layer also keeps its chunk summaries and refreshes a summary when
its chunk changes.
Generation fills blocks of 32 positions.
Inside a block, every token whose confidence is at least 0.9 is committed,
and each step commits at least the most confident token.
On RULER the masked span has the same length as the gold answer.
On LongBench each task keeps its official generation length.

The two benchmarks use different prompts.
A RULER prompt repeats the filler sentence to the target length, inserts the
task's needles into that filler, and ends with the question.
The single-needle prompt is shown below.
Multi-key multi-query asks for the codes of two named items, and variable
tracking asks for every variable assigned a given value.

\vspace{0.4em}
\noindent
\textbf{RULER.}
\par\vspace{0.3em}
\noindent
\fbox{\begin{minipage}{\dimexpr\linewidth-2\fboxsep-2\fboxrule\relax}
\small\ttfamily\raggedright
The grass is green. The sky is blue. The sun is yellow. Here we go. There and back again.\\
...\\
|One of the special magic numbers for long-context is: <code>|\\
...\\
What is the special magic number for long-context mentioned in the provided text? Answer:
\end{minipage}}

\vspace{0.8em}
\noindent
LongBench fills the official template of each task.
The multi-field English question-answering template is shown below.

\vspace{0.4em}
\noindent
\textbf{LongBench.}
\par\vspace{0.3em}
\noindent
\fbox{\begin{minipage}{\dimexpr\linewidth-2\fboxsep-2\fboxrule\relax}
\small\ttfamily\raggedright
Read the following text and answer briefly.\\[0.45em]
<context>\\[0.45em]
Now, answer the following question based on the above text, only give me the answer and do not output any other words.\\[0.45em]
Question: <question>\\
Answer:
\end{minipage}}
\vspace{0.6em}

\subsection{Training curves}
\label{sec:training_curves}

\begin{figure*}[t]
\centering
\includegraphics[width=\textwidth]{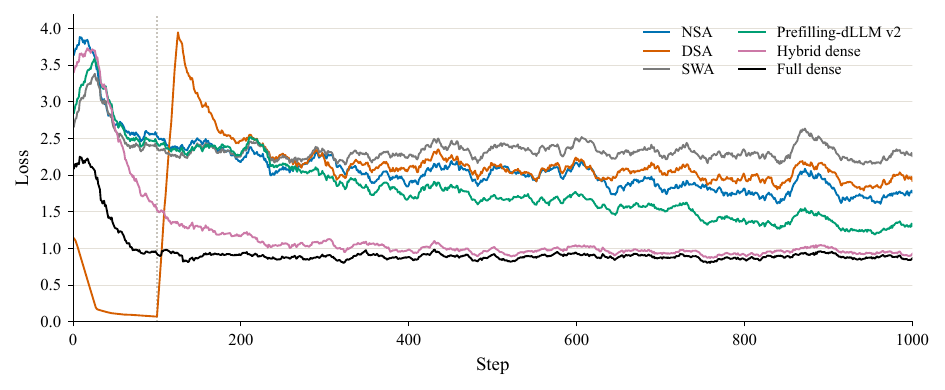}
\caption{Training loss of the continued-training runs.
Each line averages the previous 25 steps.
The dashed vertical line marks the end of the DSA indexer warmup.}
\label{fig:training_loss}
\end{figure*}

Figure~\ref{fig:training_loss} plots the continued-training objective in
\eqref{eq:sync-loss} for the six variants on one set of axes.
Each line is a trailing average of 25 steps.
DSA trains only the indexer for the first 100 steps, so its language-model
loss stays near zero until step 101, where the vertical line is drawn.

\section{Ablation study}
\label{sec:ablation}

Every run in this section is evaluated with the YaRN factor held at 8 at
every length.
The reference row in both tables is therefore the \textit{Fixed} row of \ours
in Table~\ref{tab:ruler}.

The component ablations keep the native mask and remove one piece of the full
model.
\emph{No residual} drops the rank-64 query residual.
\emph{No role offset} keeps the mask embedding and does not add the learned
role offset.
\emph{Mean summary} replaces the attention pool in \eqref{eq:summary} with the
mean of the content keys.
\emph{No entropy} removes $H(\alpha_c)$ from the chunk score.
These runs use the same continued-training recipe as \ours.
Table~\ref{tab:ablation_components} reports the four variants at 16K, 32K,
64K, and 128K.

Removing the residual is the largest drop.
The means of \emph{No residual} are 26.0, 19.1, 12.9, and 9.1, and its
single-needle accuracy falls from 63.0 at 16K to 16.0 at 128K.
The rank-64 residual is part of the query that scores the chunks.
The means of \emph{No role offset} are 40.5, 39.7, 41.3, and 37.3.
Its single-needle accuracy is 95.0, 94.0, 93.0, and 85.0, while its multi-key
accuracy is 13.5, 9.5, 13.0, and 13.5, compared with 42.0, 35.0, 27.0, and 32.5
for \ours.
Without the offset the model still finds a single needle and loses most
multi-key retrieval.
The means of \emph{No entropy} are 40.7, 38.8, 37.3, and 27.3.
Its single-needle accuracy is 98.0, 97.0, and 90.0 through 64K and falls to
58.0 at 128K, while its multi-key accuracy stays between 7.0 and 13.0.
\emph{Mean summary} is the one variant that is not below \ours at the shorter
lengths.
Its means are 56.5, 56.6, and 52.1 at 16K, 32K, and 64K, compared with 53.8,
50.7, and 48.0, and its multi-key accuracy is 50.5, 51.0, and 39.0, compared
with 42.0, 35.0, and 27.0.
At 128K its single-needle accuracy falls to 62.0, compared with 99.0, and its
mean falls to 35.9, compared with 46.5.

\begin{table*}[t]
\centering
\scriptsize
\setlength{\tabcolsep}{3pt}
\renewcommand{\arraystretch}{1.12}
\setlength{\belowcaptionskip}{8pt}
\caption{Component ablations on the three RULER tasks.
The YaRN factor is held at 8 at every length.
MK is multi-key multi-query.
Avg.\ is the mean of S-N, MK, and VT.
\ours is the full model, copied from its \textit{Fixed} row in
Table~\ref{tab:ruler}.
Each other row removes one component.}
\label{tab:ablation_components}
\resizebox{\linewidth}{!}{%
\begin{tabular}{@{}l *{16}{c}@{}}
\toprule
& \multicolumn{4}{c}{\textbf{16K}}
& \multicolumn{4}{c}{\textbf{32K}}
& \multicolumn{4}{c}{\textbf{64K}}
& \multicolumn{4}{c}{\textbf{128K}} \\
\cmidrule(lr){2-5} \cmidrule(lr){6-9} \cmidrule(lr){10-13} \cmidrule(lr){14-17}
Variant
& S-N & MK & VT & Avg.
& S-N & MK & VT & Avg.
& S-N & MK & VT & Avg.
& S-N & MK & VT & Avg. \\
\midrule
\rowcolor{mycolor!30}
\ours
& 100.0 & 42.0 & 19.3 & 53.8
& 99.0 & 35.0 & 18.0 & 50.7
& 100.0 & 27.0 & 17.0 & 48.0
& 99.0 & 32.5 & 8.0 & 46.5 \\
No residual
& 63.0 & 12.0 & 3.0 & 26.0 & 47.0 & 5.5 & 4.7 & 19.1
& 27.0 & 5.5 & 6.3 & 12.9 & 16.0 & 7.5 & 3.7 & 9.1 \\
No role offset
& 95.0 & 13.5 & 13.0 & 40.5 & 94.0 & 9.5 & 15.7 & 39.7
& 93.0 & 13.0 & 18.0 & 41.3 & 85.0 & 13.5 & 13.3 & 37.3 \\
Mean summary
& 99.0 & 50.5 & 20.0 & 56.5 & 100.0 & 51.0 & 18.7 & 56.6
& 96.0 & 39.0 & 21.3 & 52.1 & 62.0 & 29.5 & 16.3 & 35.9 \\
No entropy
& 98.0 & 13.0 & 11.0 & 40.7 & 97.0 & 7.0 & 12.3 & 38.8
& 90.0 & 10.0 & 12.0 & 37.3 & 58.0 & 12.5 & 11.3 & 27.3 \\
\bottomrule
\end{tabular}}
\end{table*}

The token ablations replace the routing mask and keep the rest of \ours.
\emph{EOS} writes the end-of-sequence token into the routing slot.
\emph{New token} uses a learned embedding in place of the native mask.
At evaluation the routing slot holds the token each model was trained with.
Table~\ref{tab:ablation_tokens} reports both at every length.
The means of \emph{EOS} are 37.5, 36.2, 31.8, and 22.8, and the means of
\emph{New token} are 38.9, 37.9, 39.9, and 30.7, compared with 53.8, 50.7,
48.0, and 46.5 for the native mask.
Multi-key retrieval is lower at every length: \emph{EOS} scores 10.5, 4.5,
7.0, and 5.0, and \emph{New token} scores 7.5, 4.0, 7.0, and 5.0, compared with
42.0, 35.0, 27.0, and 32.5.
Single-needle accuracy is within 7 points of the native mask at 16K and 32K,
where
\emph{EOS} scores 98.0 and 97.0 and \emph{New token} scores 93.0 and 94.0,
compared with 100.0 and 99.0.
At 64K and 128K \emph{EOS} scores 81.0 and 57.0 and \emph{New token} scores
93.0 and 72.0, while the native mask scores 100.0 and 99.0.
On variable tracking \emph{EOS} stays between 4.0 and 7.3, while
\emph{New token} scores 16.3, 15.7, 19.7, and 15.0, compared with 19.3, 18.0,
17.0, and 8.0.
Both replacement tokens can retrieve a single needle at the training length
and lose most multi-key retrieval at every length.
The native mask is the only slot that keeps both.
The 16K single-needle scores of \emph{EOS} and \emph{New token} differ by one
example from the clean bars of Figure~\ref{fig:mask-probe}, which come from a
separate run of the probe.

\begin{table*}[t]
\centering
\scriptsize
\setlength{\tabcolsep}{3pt}
\renewcommand{\arraystretch}{1.12}
\setlength{\belowcaptionskip}{8pt}
\caption{Routing-token ablations on the three RULER tasks.
The YaRN factor is held at 8 at every length.
MK is multi-key multi-query.
Avg.\ is the mean of S-N, MK, and VT.
\emph{Mask} is the native mask used by \ours, copied from its
\textit{Fixed} row in Table~\ref{tab:ruler}.
\emph{EOS} and \emph{New token} replace the mask in the routing slot, both
in training and at evaluation.}
\label{tab:ablation_tokens}
\resizebox{\linewidth}{!}{%
\begin{tabular}{@{}l *{16}{c}@{}}
\toprule
& \multicolumn{4}{c}{\textbf{16K}}
& \multicolumn{4}{c}{\textbf{32K}}
& \multicolumn{4}{c}{\textbf{64K}}
& \multicolumn{4}{c}{\textbf{128K}} \\
\cmidrule(lr){2-5} \cmidrule(lr){6-9} \cmidrule(lr){10-13} \cmidrule(lr){14-17}
Variant
& S-N & MK & VT & Avg.
& S-N & MK & VT & Avg.
& S-N & MK & VT & Avg.
& S-N & MK & VT & Avg. \\
\midrule
\rowcolor{mycolor!30}
Mask
& 100.0 & 42.0 & 19.3 & 53.8
& 99.0 & 35.0 & 18.0 & 50.7
& 100.0 & 27.0 & 17.0 & 48.0
& 99.0 & 32.5 & 8.0 & 46.5 \\
EOS
& 98.0 & 10.5 & 4.0 & 37.5 & 97.0 & 4.5 & 7.0 & 36.2
& 81.0 & 7.0 & 7.3 & 31.8 & 57.0 & 5.0 & 6.3 & 22.8 \\
New token
& 93.0 & 7.5 & 16.3 & 38.9 & 94.0 & 4.0 & 15.7 & 37.9
& 93.0 & 7.0 & 19.7 & 39.9 & 72.0 & 5.0 & 15.0 & 30.7 \\
\bottomrule
\end{tabular}}
\end{table*}

\end{document}